\documentclass{article}

\usepackage{fancybox}
\usepackage{listings}
\usepackage{float}
\usepackage{footnote}
\usepackage{hyperref}
\usepackage{placeins}
\usepackage{booktabs} 	
\usepackage{tabularx}
\usepackage{array}
\usepackage{subfigure}
\usepackage{tikz}
\usepackage{fancyvrb}

\usepackage{subfigure}
\usepackage{caption}
\usepackage{multirow}
\usepackage{verbatim}
\usepackage{moreverb}

\usepackage{graphicx}
\usepackage{hyperref}
\usepackage{listings}
\usepackage{float}
\usepackage{graphics}
\usepackage{algorithm2e}
\usepackage{paralist}
\usepackage{amssymb}
\usepackage{amsmath}
\usepackage{listings}
\usepackage{setspace}
\usepackage{caption}
\usepackage[T1]{fontenc}
\usepackage{textcomp}
\usepackage[utf8]{inputenc}
\usepackage{csquotes}
\usepackage{newunicodechar} 
\usepackage{lstautogobble}
\usepackage{times}
\usepackage{latexsym}
\usepackage{lipsum}
\usepackage{pgf}
\usepackage{fancyvrb,xcolor}
\usepackage{framed,color}
\usepackage{float}
\usepackage{authblk}

\usepackage{caption}

\newtheorem{req}{Requirement}

\usepackage{verbatim}

\usepackage{todonotes} 

\usepackage{etoolbox}
\usepackage{fancybox}
\usepackage{listings}
\usepackage{enumerate}
\usepackage{hyperref}
\usepackage{float}

\definecolor{mygreen}{HTML}{00CC00}

\newtheorem{definition}{Definition}

\newcommand{\urlfn}[1]{\footnote{\scriptsize\url{#1}}}

  \renewenvironment{thebibliography}[1]{%
    \begin{oldthebibliography}{#1}%
      \setlength{\parskip}{0ex}%
      \setlength{\itemsep}{0ex}%
  }%
  {%
    \end{oldthebibliography}%
  }

\usepackage{color}
\definecolor{olivegreen}{rgb}{0.2,0.8,0.5}
\definecolor{grey}{rgb}{0.5,0.5,0.5}
\lstdefinelanguage{ttl}{
sensitive=true,
morecomment=[l][\color{grey}]{@},
morecomment=[l][\color{olivegreen}]{\#},
morestring=[b][\color{blue}]\",
keywordstyle=\color{cyan},
morekeywords={version,owl,rdf,rdfs,xml,xsd,dbpedia,dbo,str,sso,scms,fr,ld}
}

\definecolor{shadecolor}{rgb}{.9, .9, .9}
\newenvironment{Code}%
   {\snugshade \verbatim \scriptsize}%
   {\endverbatim\endsnugshade}

\begin{document}

\title{CEVO: Comprehensive EVent Ontology 
\\ Enhancing Cognitive Annotation on Relations}

\author[1]{Saeedeh Shekarpour}
\author[2]{Faisal Alshargi}
\author[3]{Krishnaprasad Thirunarayan}
\author[3]{Valerie L. Shalin}
\author[3]{Amit Sheth}
\affil[1]{University of Dayton, USA, sshekarpour1@udayton.edu}
\affil[2]{University of Leipzig, Germany, alshargi@informatik.uni-leipzig.de}
\affil[3]{DKno.e.sis Center,  \{tkprasad, valerie, amit\}@knoesis.org}

\maketitle

\begin{abstract}
While the general analysis of named entities has received substantial research attention on unstructured as well as structured data, the analysis of relations among named entities has received limited focus.
In fact, a review of the literature revealed a deficiency in research on the abstract conceptualization required to organize relations.  
We believe that such an abstract conceptualization can benefit various communities and applications 
such as natural language processing, information extraction, machine learning, and ontology engineering.
In this paper, we present Comprehensive EVent Ontology (CEVO), built on 
Levin's conceptual hierarchy of English verbs that categorizes verbs with shared meaning, and syntactic behavior. 
We present the fundamental concepts and requirements for this ontology.
Furthermore, we present three use cases employing the CEVO ontology on 
annotation tasks: (i) annotating relations in plain text, (ii) annotating ontological properties, and (iii) linking textual relations to ontological properties.
These use-cases demonstrate the benefits of using CEVO for annotation: (i) annotating English verbs from an abstract conceptualization, (ii) playing the role of an upper ontology for organizing ontological properties, and
(iii) facilitating the annotation of text relations using any underlying vocabulary.
This resource is available at \url{https://shekarpour.github.io/cevo.io/} using \url{https://w3id.org/cevo} namespace.

\end{abstract}





\

\section{Introduction}
While the size of data on the Web is growing dramatically in both structured and unstructured, still a high proportion of the Web remains unstructured (i.e., textual data) such as social network feeds, blogs, news, logs.  
However, the size of structured data is also significant; e.g., so far more than 130 billion triples are published on Linked Open Data from over 9,960 datasets\footnote{observed on September 30th 2018 at \url{http://lodstats.aksw.org/}}. These large datasets subscribe to ad hoc, independently created and diverse lexicons, nomenclatures, vocabularies, taxonomies, and ontologies.
Although this diversity provides flexibility, it complicates both reuse of ontologies and interlinking of datasets.

So far, named entity recognition and entity linking to the background knowledge base have received substantial attention in research studies. However, the analysis of relations over named entities has not. 
Although cognitive scientists e.g., Doumas and Humel, 2005  \cite{doumas2005approaches,doumas2005symbolic} suggest that relational content is key to reasoning, a review of the literature on unstructured as well as structured data revealed a deficiency in research on the abstract conceptualization required to organize relations.
Observing deficiencies in (i) \emph{relation extraction} from text, (ii) \emph{contextual equivalencing of relations}, and (iii) dealing with the \emph{diversity of ontologies} motivated us to investigate an abstract conceptualization of relations.

We found an answer to our struggle in the organized lexicon and knowledge base assembled by the Stanford linguist Beth Levin in \cite{levin_english_1993}. While Levin relies on Schank's conceptual dependency theory \cite{schank_conceptual_1972,Sch75} to organize this knowledge base, the key here is the psychologically principled inventory of English verbs aligned with the knowledge base. 
Classes in this knowledge base identify sets of semantically coherent verbs with corresponding syntactic properties.
For example, the \emph{communication} class refers to verbs causing a communication activity.
A subcategory of this class is the category of verbs transferring a message/idea (i.e. shared meaning), such as \emph{announce}, \emph{say}, and \emph{mention}. In addition, they share the same syntactic behaviour such as \texttt{NP1 VP NP2} or \texttt{NP1 VP NP2 to NP3} (\texttt{NP} is noun phrase and \texttt{VP} is verb phrase). 
For example, \texttt{Jack says greetings to Sarah} is following the syntactic pattern \texttt{NP1 VP NP2 to NP3}.
The semantic behavior is \emph{communication} activity between two entities (i.e., Jack and Sarah) with the transferred message ``greetings".
This conceptualization provides more than 230 classes for over 3,000 English verbs.
Using Levin's work, we build an event ontology and lexicon called Comprehensive EVent Ontology (\textbf{CEVO})\footnote{CEVO namespace: \url{https://w3id.org/cevo/}}.
CEVO is designed to recognize and equate relations from both textual data sources as well as knowledge bases.
Such an abstract conceptualization benefits many applications, 
such as natural language processing, information extraction, ontology engineering, and machine learning.

\begin{figure}[hpbt]
\centering
\includegraphics[width=0.8\columnwidth]{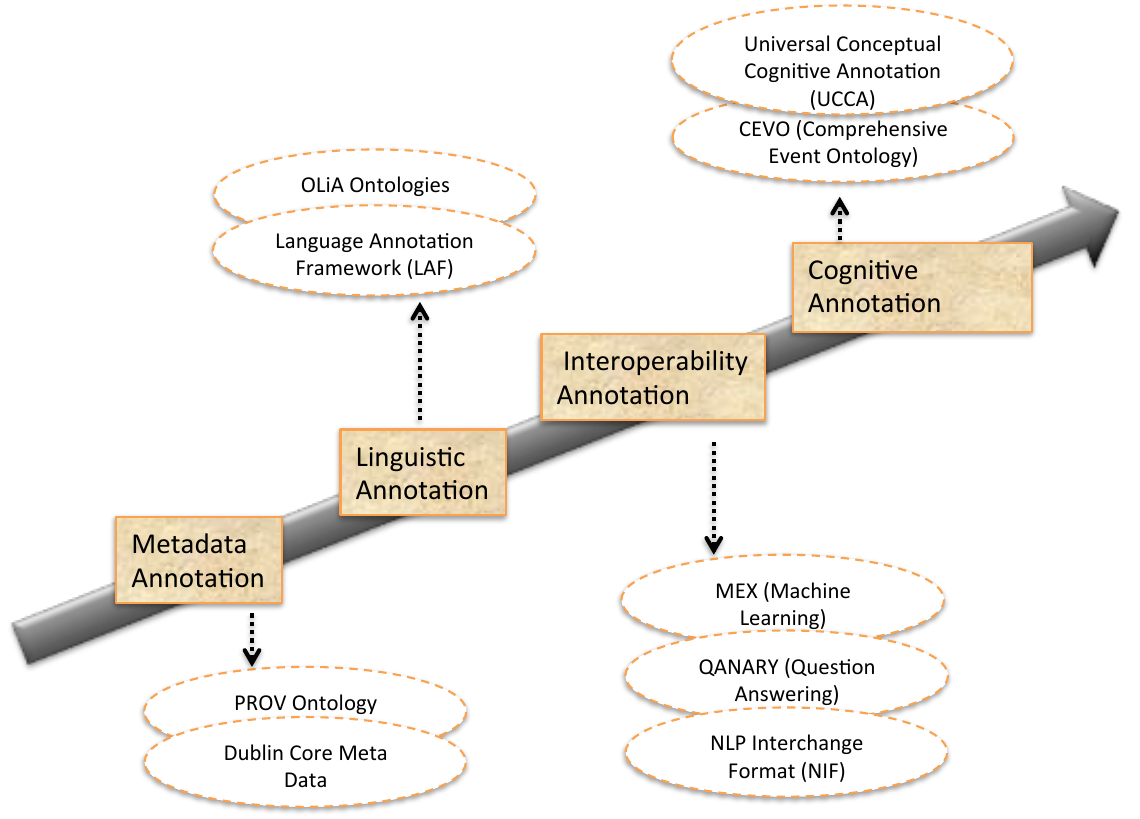}
\caption{Evolution of Abstraction Level of Vocabularies and Ontologies.}
\label{fig:evolution}
\end{figure}

Figure \ref{fig:evolution} represents the evolution of ontologies as well as vocabularies regarding the \emph{level of abstraction}. 
The early generation of vocabularies was primarily for annotating datasets (i.e., metadata) or describing a domain.
The next generation of vocabularies have supported interoperability requirements and hence are created from a higher level of abstraction.
The subsequent generation of ontologies has cognitive applicability and therefore has the highest level of abstraction.
CEVO provides a high abstraction for annotating events.

This paper is organized as follows. 
Related work is presented in Section \ref{sec:relatedwork}.
Section \ref{sec:problemStatement} presents the deficiencies that motivated us to develop the CEVO ontology.
Section \ref{sec:requirements} discusses fundamental requirements for designing the CEVO ontology.
The principles and considerations that Beth Levin has taken into account for categorizing English verbs 
are presented in Section \ref{sec:levinPrinciples}.
The main concepts of CEVO are introduced in Section \ref{sec:CEVO}. 
We discuss three use cases employing CEVO for annotation tasks in Section \ref{sec:usecases}.
We close with the conclusion and future work in Section \ref{sec:conclusionandfuturework}.

\section{Related Work}
\label{sec:relatedwork}

We first review the definition of \emph{event} which has been given previously in different ontologies and then present several existing ontologies that facilitate the annotation task and interoperability among various components.
\paragraph{\textbf{Event Classification}} \textbf{LODE:} An ontology for Linking Open Descriptions of Events\footnote{\url{http://linkedevents.org/ontology/}} defines a single generic concept of event as `Something that happened',
e.g., reported in a news article or with historical significance.
This is a generic definition and does not specify the various types of events necessary for subsequent inference.
\textbf{Schema.org}\footnote{\url{http://schema.org}} introduces a similar generic concept of event\footnote{\url{http://schema.org/Event}} that additionally considers temporal as well as location aspects and additionally provides a limited hierarchy.
This hierarchy  introduces types of events such as business events, sale events, and social events.
Similarly, the \textbf{DBpedia ontology}\footnote{\url{http://wiki.dbpedia.org/services-resources/ontology}} defines the generic concept of event with a hierarchy which is broader, including lifecycle events (e.g. birth, death), natural events (e.g. earthquake, stormsurge), and  societal events (e.g. concert, election).
However, the state-of-the-art does not provide a detailed specification about various types of events.
Having such specification is of high importance because in specific domains such as news domain it is required to differentiate between various types of events.
\emph{Event classification} task is necessary in case the employed background data model considers the specific type of events as part of event annotation. In this case, event phrases have to be labeled by specific types of events using multi-class classifier trained for distinguishing the specific type of a given event. For example, \autoref{tab:tweetsamples} shows samples of news headline annotated based on their major event (column 2) like the tweets no.2, no.5, no.8 of \autoref{tab:tweetsamples} having the specific type \textit{``meet''}.
To the best of our knowledge, CEVO is the first event ontology that provides a fine-grained abstract conceptualization of events.
In fact, CEVO captures the whole of the hierarchy of Levin 's classification as the hierarchy representing specific events. Thus, it specifies 230 events.

\begin{table}[hptb]
	\centering
\begin{tiny}
\begin{tabular}{ l|l}
\hline
  \textbf{Event Type}		& \textbf{News Headlines Tweets} 	   \\ \hline

	communication	&  	no1.  Michelle Obama tells \#SXSW crowd: I will not run for president  \\
meet	&	no2. Instagram CEO meets with @Pontifex to discuss "the power of images to unite people" 	 	 	 \\  	
 murder (killing)	&	 no3. Chemical accident in Bangkok bank kills eight people 	 		 \\ 	 \hline

	communication 		&  no4. State elections were "difficult day," German Chancellor Angela Merkel says  	  \\ 
   meet 			 & no5. Pope Francis visits Cuba and Mexico	 \\ 	
  murder (killing)		 & no6. Storms kill at least three in Virginia	  \\	\hline 
						 		
  communication    &    no7. Obama and Justin Trudeau announce efforts to fight climate change		\\	
  meet	&   no8. Pope to meet leader of Russian Orthodox Church for first time in nearly  \\ 	
  murder (killing) 	&	no9. 2 air force pilots from United Arab Emirates  killed when warplane crashed over Yemen\\  \hline

\end{tabular}
\end{tiny}
\caption{ Samples of news headlines from different publishers on Twitter.}
\label{tab:tweetsamples}
\end{table}

\paragraph{\textbf{Interoperability Challenge}} the interoperability among heterogeneous datasets, schemas, various tools, and dependent components is an important issue. So, recently, a number of linguistic ontologies as well as annotation tools for related purposes have emerged.
The Ontologies of Linguistic Annotation (OLiA) \cite{OLIA1,Olia2} provides annotation tag sets using syntactical and morphological perspectives.
OLiA covers over 110 OWL ontologies for over 34 tag sets in 69 different languages.
 Thus, all NLP tools can leverage OLiA's tag sets for annotating the output.
Among the ontologies introduced for promoting interoperability between tools, services, and components, we mention two recent ones: 
 
 \begin{enumerate}
 \item 
 NLP Interchange Format (NIF)\footnote{\url{http://persistence.uni-leipzig.org/nlp2rdf/}} \cite{NIF} provides a vocabulary for interoperability between Natural Language Tools (NLP).
NIF allows tools to exchange annotations for any part of text using three layers: (i) \emph{Structural layer:} This describes URI scheme for identifying any part of a document; thus, making each piece of text dereferenceable. 
(ii) \emph{Conceptual layer:} The NIF Core Ontology\footnote{\url{http://persistence.uni-leipzig.org/nlp2rdf/ontologies/nif-core/nif-core.html\#}} 
 describes classes and properties for providing relations between tokens and documents. 
The core class is \texttt{nif:String}, which points to any word mentioned in Unicode characters.
(iii) \emph{Access layer:} NIF-aware applications publish their output using the NIF format via REST APIs.

\item
qa vocabulary\footnote{\url{https://github.com/WDAqua/QAOntology}} \cite{QANARY1,QANARY2} is intended to facilitate interoperability among components of question answering systems.
Currently, it defines a two-layered ontology: a \emph{abstract layer} describing the generic functionalities of each component and a \emph{binding layer} enabling binding output of each element to the abstract layer.
\end{enumerate}

CEVO connects the conceptualization of events to a large lexicon of English verbs, i.e., 3000 verbs. Thus, it can provide an abstract annotation on relations.
Please be noted that this knowledge base is different from lexicons such as \emph{lemon}\footnote{\url{http://lemon-model.net/learn/5mins.html}} \cite{lemon2011,lemon2013} which shares purely terminological and lexicological resources on the Semantic Web. 
Potentially, CEVO conceptualization will play a significant role in overcoming 
the existing challenges of (i) tagging relations, (ii)  linking relations, and (iii) ontology alignment.


\section{Problem Statement}
\label{sec:problemStatement}

The CEVO ontology compensates for pervasive deficiencies in the abstract conceptualization of relations.
In the following, we mention the three well-known deficiencies in annotating relations or ontological properties.


\begin{enumerate}[(I)]
\item {\textbf{ Relation  Extraction}}: Decades of research in the field of information extraction has resulted in tools that successfully recognize and annotate Named Entities (NE) and link them to entities available in a background knowledge base (e.g., \cite{StanfordNE,tagme2,DBPediaspotlight}). In contrast, recognizing, tagging, and linking relationships among entities has received limited attention.
\item {\textbf{Contextual Equivalence of Relations}}:
Relations embedded in plain text can be expressed in various ways either explicitly or implicitly.
Explicit relations often appear as a verb phrase and implicit ones are usually hidden or embedded in other phrases, e.g., an adjective phrase such as "European Countries" (countries located in Europe).
Moreover, a single explicit relation can be expressed using several distinct verbs. E.g.,  consider the two sentences `Jack visits Sara', and `Jack consults Sara'. 
In both of these cases, the \emph{abstract event} of \emph{meeting} is taking place using two different verbs, \emph{`visit'} and \emph{`consult'}.  Nevertheless,  these two verbs do not have a simple synonymous relationship recoverable  by inference using lexicons such as WordNet. These verbs convey the same event in a specific context.

\item {\textbf{Diversity in conceptualization}}: 
Each ontology is created based upon a specific interpretation of a domain. 
This strategy yields various ontologies used by different users or communities.
While this heterogeneity provides flexibility, it complicates reusability and introduces integration challenges. Linking the mention of either entity or relation from plain text to the corresponding background knowledge base is ontology-specific.

\end{enumerate}

\section{Requirements for CEVO}
\label{sec:requirements}
On the basis of these observations, we derive a series of requirements for a cognitive ontology for annotating relations on both structured as well as unstructured data. The coarse-grained requirements are listed below.

\begin{req}[Relation Tagging on Textual Data]
Similar to tagging Named Entities in plain text, each mention of a relation must be recognized, normalized, and tagged. 
Thus, a tags set is required for distinguishing relations.
\end{req}

\begin{req}[Relation Linking ]
Beyond recognizing and tagging mentions of relations in plain text, it is necessary to link textual relations to ontological properties. 
To do that, having an upper ontology that can be used for annotating both ontological properties and textual relations is required.
\end{req}




\begin{req}[Integration and Alignment of Properties]
The variety of ways to conceptualize a domain results in different ontologies.
Certainly, overlaps between ontologies require alignment or integration. Thus, annotating ontologies based on an upper ontology which has a higher abstraction
can help in integrating and aligning ontologies.
\end{req}

\begin{req}[Reusability]
One of the main obstacles for the reuse of ontologies is the additional effort required for interpreting the conceptualization represented.
Providing annotations based on a cognitive conceptualization which is from a higher abstraction level indeed boosts reusability because it is not involved in domain-specific considerations.
\end{req}

\begin{req}[Simplicity]
Since we desire CEVO to be widely adopted and reused, the captured cognitive conceptualization has to be as simple as possible to minimize integration and adoption efforts.
\end{req}


To the best of our knowledge,  there is no existing ontology on Linked Data that fulfills the above requirements. 
To this end, we present CEVO (a cognitive event ontology), built upon \cite{levin_english_1993}. 
We contend that CEVO fulfills an abstract conceptualization of relations.
In the following section, we discuss principles behind this conceptual hierarchy.

\begin{figure*}[!hptb]
\centering
\includegraphics[width=0.7\textwidth]{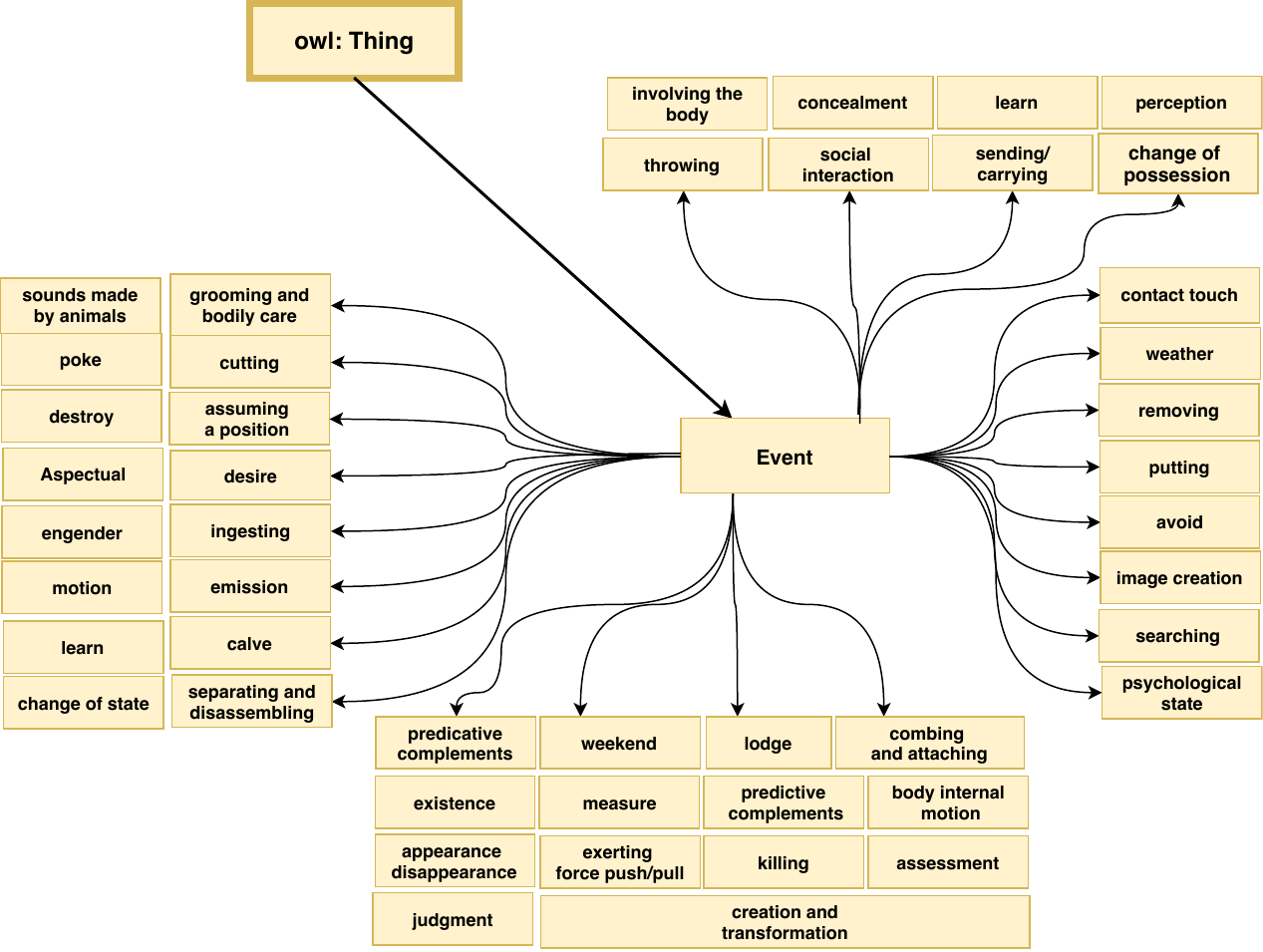}
\caption{First level of event hierarchy}
\label{fig:eventHierarchy}
\end{figure*}

\section{Levin Conceptual Hierarchy}
\label{sec:levinPrinciples}
 
\begin{figure}[!hptb]
\centering
\includegraphics[width=0.9\columnwidth]{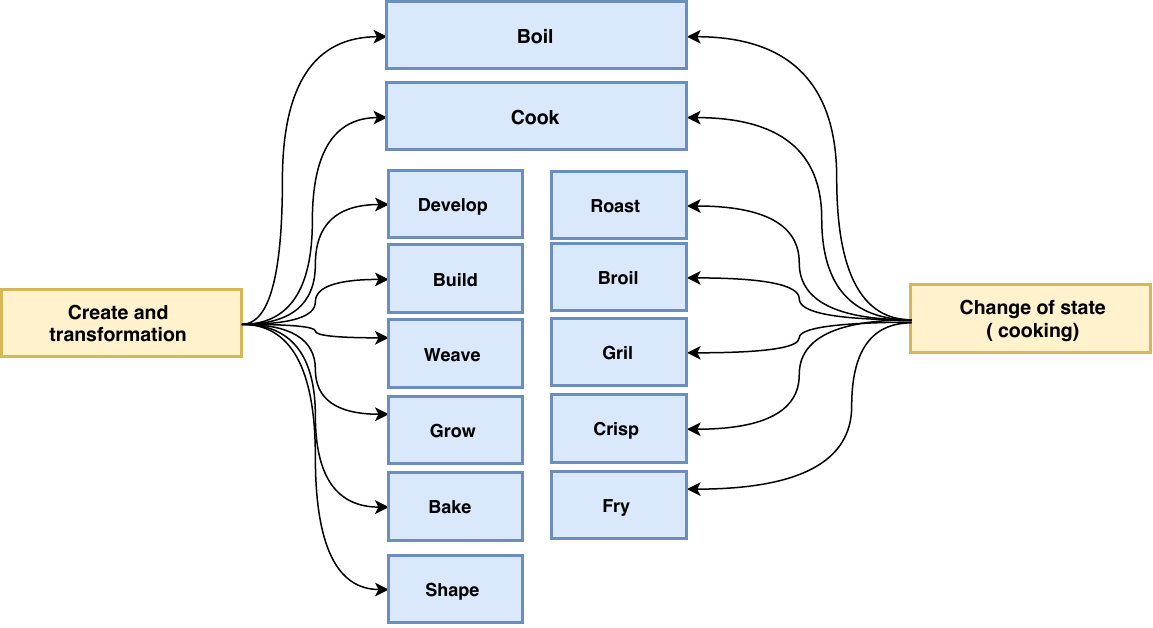}
\caption{Two classes of Levin categorization with samples of member verbs: (i) Creation and transformation class and (ii) Change of the state class}
\label{fig:classes}
\end{figure}

The entries of Levin's lexical knowledge base are verb classes. 
For example, Figure \ref{fig:classes} illustrates two distinct English verb classes (1) \emph{transformation and creation} and (2) \emph{change of the state}  both of which subsume several verbs.
The members of each class (i.e. English verbs) have\ two characteristics:
(i) semantic coherence and (ii) shared syntactic behavior. These characteristics are described as follows:

\emph{(i) Semantic Coherence:} Each class of verbs has a unique set of properties that shape the meaning of the member verbs (i.e. shared meaning).
In fact, the conjunction of properties provides a distinctive meaning for each class. A single meaning property might be attributed to several classes depending upon context.  Moreover, an individual verb can belong to multiple classes, creating a graph instead of a tree. 
For example, the class of \emph{Creation and Transformation} (shown in Figure \ref{fig:classes}) refers to a class of verbs causing alternation. 
This class contains both transitive verbs (an agent creates an entity) and intransitive verbs (describing transformation of an entity).
A subclass of this class is the \emph{Build} class with the shared properties: (1) material/product alternation, (2) total transformation alternation, (3) unspecified object alternation, (4) benefactive alternation, (5) causative alternation, (6) raw material subject alternation, and (7) sum of money subject alternation; whereas another subclass namely \emph{Grow} class shares only three properties: (1) material/product alternation, (2) total transformation alternation, and (3) causative alternation.

An English verb might belong to two distinct classes. 
For example, the verbs \emph{`cook'} and \emph{`boil'}  belong to two distinct classes: (a) creation and transformation and (b) change of state.
These two classes are represented in Figure \ref{fig:classes}.
Thus, depending on the \emph{context}, the appropriate class is distinguished. 

\emph{(ii) Shared syntactic behavior:} Meaning influences the syntactic behavior of a verb in terms of its expression and the interpretation of its arguments. Verbs with shared meaning exhibit similar syntactic behavior.
For example, \texttt{NP1 VP from NP2 to NP3} (e.g., Jack flew from London to New York) is a syntactic pattern that might be shared between verbs categorized in a class.

\section{CEVO: Comprehensive EVent Ontology}
\label{sec:CEVO}
In this section, we describe the main concepts introduced in CEVO in the schema as well as instance level.
We use the prefix \texttt{cevo: <https://w3id.org/cevo/>} in the following.

\subsection{CEVO Schema} 
The top class of CEVO is the class of generic \emph{Event}, which is the superclass of all specific events.
The generic Event class is formally defined as follows:

\begin{definition} [Class of generic events]
The generic \texttt{cevo:Event} is an \texttt{owl:Class} and refers to `occurrence of anything'. It is generally the superclass of any specific type of event.
\end{definition}

\begin{Code}
cevo:Event       a             owl:Class                  ;  
                 rdfs:label    `generic event'            ;
                 rdfs:comment  `something that happens'   .
\end{Code}

\begin{figure}[!hptb]
\centering
\includegraphics[width=0.9\columnwidth]{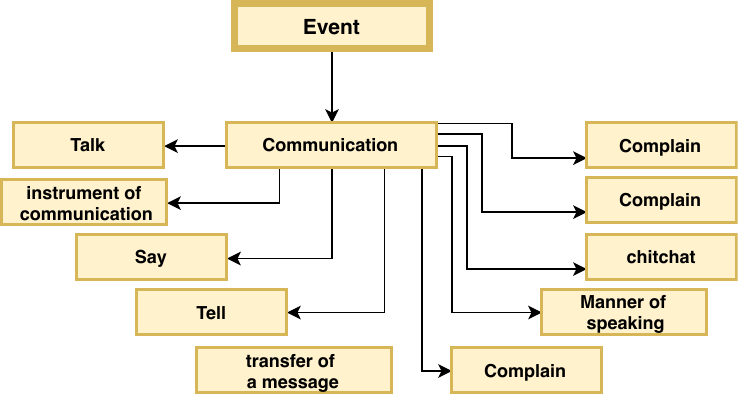}
\caption{Hierarchy of \texttt{cevo:Communication} event.}
\label{fig:communicationEvent}
\end{figure}

In CEVO, the Levin conceptual hierarchy is incorporated under the generic \emph{Event} class. 
Figure \ref{fig:eventHierarchy} illustrates the first level of Levin 's hierarchy.
In other words, any class provided for a set of English verbs revealing a specific event is considered as an 
\texttt{owl:Class} as:

\begin{definition} [Class of `X' events]
\texttt{cevo:X} event refers to an specific event and is subclass of the class \texttt{cevo:Event}. Conceptually, it refers to a specific type of event that is associated with an English verb category sharing a common behavior or meaning. \end{definition}
For instance, the class \texttt{cevo:Communication} given below is defined as a subclass of \texttt{cevo:Event}. This class refers to the occurrence of any activity for communicating or transferring a message/idea.
Figure \ref{fig:communicationEvent} represents the hierarchy of the communication event in CEVO where \texttt{cevo:Communication} is divided into eight subclasses. For example, \texttt{cevo:Complain} specifies events showing the speakers's attitude or feeling towards what is said in addition to the communicating activity. 

\begin{Code}


cevo:Communication   a                   owl:Class     ; 
                     rdfs:subClassOf      cevo:Event   ;  
                     rdfs:label          `communication ;
                     rdfs:comment        `communication and
                     	                  transfer of idea' .
\end{Code}

The next main class is \texttt{cevo:Verb} refers to words with the part-of-speech \emph{verb}.
This class is equivalent to the class \emph{main verb} of \emph{OLiA} ontology \cite{OLIA1,Olia2}\footnote{OLiA Ontology is available at \url{http://nachhalt.sfb632.uni-potsdam.de/owl/}} \footnote{Documentation of OLiA is available at \url{http://nachhalt.sfb632.uni-potsdam.de/owl/olia.owl}} which is an annotation model based on morphology.

\begin{Code}
cevo:Verb          a                        owl:Class     ;
                   owl:equivalentClass      OLiA:MainVerb . 
\end{Code}

\subsection{CEVO Instance Level (Verb Individuals)}
\begin{figure}[!hptb]
\centering
\includegraphics[width=0.9\columnwidth,]{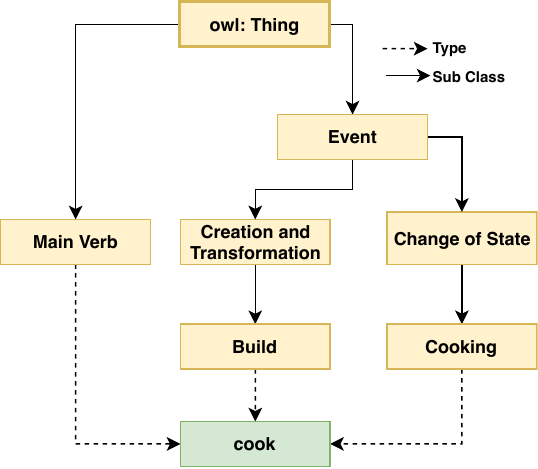}
\caption{Classes associated with the verb \texttt{cevov:cook}.}
\label{fig:cookVerb}
\end{figure}

So far, we described the schema level classes containing two major classes \texttt{cevo:Event} and \texttt{cevo:Verb}; the next important step is to map each individual English verb to the corresponding event class.
Because any \textbf{appearance of a verb} shows the likelihood of occurrence one or multiple events. 
In other words, verbs expose occurrences of events.
Thus, each individual verb has a dual sort of \texttt{rdf:type} (i) the class \texttt{cevo: Verb} meaning this individual linguistically is a verb and (ii) \texttt{cevo:Event} meaning this verb manifests occurrence of an event.
We use the prefix \texttt{cevov: <https://w3id.org/cevo/verb/>} for referring verb individuals.
Thus, we instantiate each English verb at the instance level and type that primarily as \texttt{cevo:Verb} and then map it to the associated event(s).
In the following, the two English verbs \emph{say} and \emph{cook} are instantiated. 
They are primarily typed as \texttt{cevo:Verb}, and furthermore, they are typed to their corresponding event classes, respectively \texttt{cevo:Communication} and \texttt{cevo:Creation and Transformation} events.
In fact, by specifying the type of a verb as \texttt{cevo:Verb}, explicitly, the syntactic role of that verb in the English language is determined and by associating the relevant events to a verb, domain-specific semantic roles of that verb are determined.
Please note that, each individual verb can be associated with several event classes. 
For instance, the verb \texttt{cevov:cook}, in addition to the event \texttt{cevo:Creation and Transformation}, is also associated with the event \texttt{cevo:Change of the State} (shown in Figure \ref{fig:cookVerb}).

\begin{Code}
(a) cevov:say   rdf:type   cevo:MainVerb                .
(a) cevov:say   rdf:type   cevo:Communication           .
(b) cevov:cook  rdf:type   cevo:MainVerb                .
(b) cevov:cook  rdf:type   cevo:Creation_Transformation .
(b) cevov:cook  rdf:type   cevo:Change_of_the_state     . 
(b) cevov:cook  rdf:type   cevo:Cooking                 . 
(b) cevov:cook  rdf:type   cevo:Build                   . 
\end{Code}

\begin{figure}[!hp]
\centering
\includegraphics[width=0.9\columnwidth,]{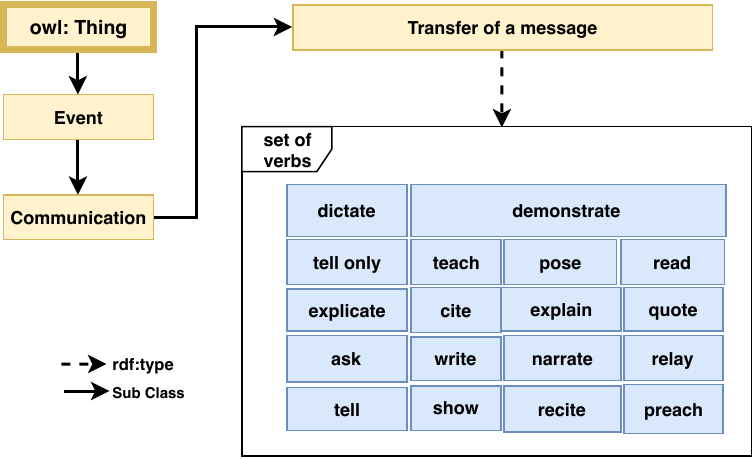}
\caption{Verbs associated with the event} \texttt{cevo:Transfer of a message.}
\label{fig:transferMessageVerbs}
\end{figure}

On the other hand, a group of verbs reveals the occurrence of an event. 
For example, Figure \ref{fig:transferMessageVerbs} represents all the verbs which may cause a \emph{Transferring a message} event (a subclass of  \texttt{cevo:Communication}).

\subsection{Querying CEVO}

A typical concern is how we can obtain the list of events related to a particular verb. 
Let us assume that the input verb \texttt{cook} is given.
The following SPARQL query retrieves the event classes associated with the input verb \texttt{cook}.
\begin{Code}
SELECT distinct ?v ?event
WHERE { 
?v rdf:type cevo:Verb .
?v rdf:type ?event.
?v rdfs:label ?vlabel.
Filter (regex(?vlabel,'cook')).
Filter (?event != cevo:Verb).
Filter (?event != owl:NamedIndividual ).}\end{Code}


\section{Use Cases}
\label{sec:usecases}
We present three use cases employing CEVO for annotating textual relations and ontological relations. 
\subsection{Use Case 1: Annotating Relations in Text using CEVO}
\begin{figure}[h]
\centering
\includegraphics[width=\columnwidth]{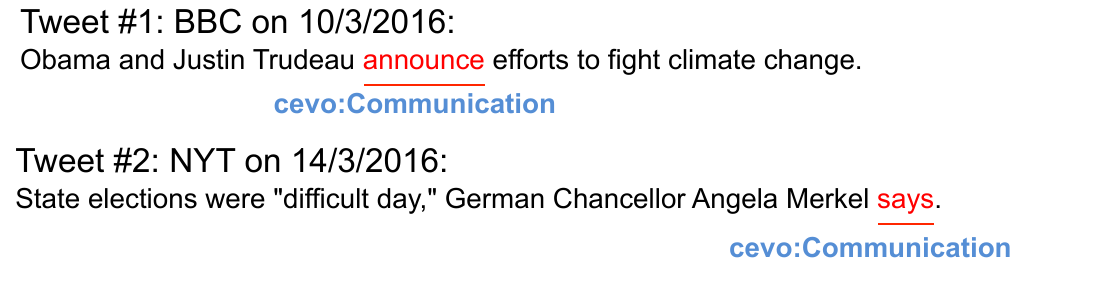}
\caption{Annotating relations expressed in headlines from news published on Twitter using CEVO. 
Both of these tweets obtained similar tags because, in both,
an event of `transferring a message'  appears. }
\label{fig:tweet}
\end{figure}

Here, we show the applicability of CEVO for annotating relations in plain text. 
Figure \ref{fig:tweet} shows two headlines from news on Twitter. The first tweet was published by the BBC and the second one was published by the New York Times.
Tweet \#1 is headed by the verb \texttt{announce} and the tweet \#2 is headed by the verb \texttt{say}. 
Both of these tweets have a similar meaning in the sense that a \emph{message is transferred}.  
Annotating these two tweets via CEVO uses the same tag \emph{communication} for both of these verbs,
whereas they do no hold any lexical relation such as synonymy.

For specificity, we first annotate the two strings \texttt{announce} and \texttt{say} using NIF vocabulary.
 
\begin{Code}
:tweetID#char=26,33     a                  NIF:String    ;
                        NIF:beginIndex     26            ;
                        NIF:endIndex       33            ;
                        NIF:anchorOf       "announce"    ;
                        NIF:oliaCategory   Olia:MainVerb .         
                              
:tweetID#char=71,74     a                  NIF:String    ;
                        NIF:beginIndex     71            ;
                        NIF:endIndex       74            ;
                        NIF:anchorOf       "says"    ;
                        NIF:oliaCategory   Olia:MainVerb .         
          
\end{Code}                              

We enhance the annotations  with the relevant event type using CEVO as follows:
\begin{Code}
tweetID#char=26,33      a         cevo:Communication  .   

tweetID#char=71,74      a         cevo:Communication  .
\end{Code}
\subsection{Use Case 2: Annotating Ontological Properties}
CEVO can play the role of upper ontology, thus it can be utilized for annotating properties of any ontology.
One way of providing such an annotation is using the Web Annotation Data Model\footnote{ W3C Recommendation since 23 February 2017, \url{http://www.w3.org/TR/annotation-model}} (WADM), which is a framework for expressing annotations.
A WADM annotation has two elements (i) a target which indicates the resource being annotated, and (ii) a body which indicates the description. 
Annotating properties of various ontologies with CEVO addresses integration and alignment problems.
Assume that we have the property \\\texttt{ dbp:spouse}\footnote{We use the prefix \texttt{dbo: <https://dbpedia.org/ontology/>} and the prefix \texttt{dbp: <https://dbpedia.org/property/>.}}  from DBpedia ontology, which represents the relation of \emph{marrying} that is semantically related to the class \texttt{ cevo:Amalgamate} (please be noted that the verb \texttt{cevov:marry} belongs to the the class \texttt{ cevo:Amalgamate}).
 The annotation of this property is presented in Turtle syntax using the WADM framework as follows:

\begin{Code}
example:annotation1    a                 oa:Annotation        ;
                       oa:hasTarget      dbp:spouse           ;
                       oa:hasBody        cevo:Amalgamate      .
\end{Code}

\subsection{Use case 3: Linking Relations}
CEVO  facilitates linking occurrences of relations in plain text to ontological properties.
For example,
on the March 4th 2016, the BBC published this headline: \texttt{Rupert Murdoch and Jerry Hall marry}.
The embedded relation in this part of text is \texttt{marry}. This relation is  annotated as \texttt{cevo:Amalgamate} employing CEVO ontology.
We show this annotation using the NIF vocabulary below.


\begin{Code}
:headline#char=31,35    a                  nif:String    ;
                        nif:beginIndex     31            ;
                        nif:endIndex       35            ;
                        nif:anchorOf       `marry'       ;
                        nif:oliaCategory   Olia:MainVerb .  
                               
                        a             cevo:Amalgamate    .
\end{Code}                              

\texttt{:headline\#char=31,35} is the assigned URI for the verb \texttt{marry} in the headline mentioned.
By taking into account the previous annotation for \texttt{dbp:spouse} property, we are now empowered to link the \texttt{marry} relation in the headline directly to the property \texttt{dbp:spouse} due to the similar tag of the CEVO event. This  annotation is represented using WADM as follows:


\begin{Code}
:annotation3   a             oa:Annotation             ;
       	       oa:hasTarget  :headline#char=31,35      ;
       	       oa:hasBody    dbo:spouse                . 
\end{Code}

Thus, using SPARQL query, we can link a textual relation to the appropriate ontological relation based on their common CEVO annotations.
For example, the verb \texttt{marry} is linked to the property \texttt{dbp:spouse} as follows:

\begin{Code}
:headline#char=31,35    itsrdf:taIdentRef   dbp:spouse .
\end{Code}                                                       


\section{Conclusions and Future Directions}
\label{sec:conclusionandfuturework}
In this paper, we introduced an event ontology called \emph{CEVO} (available at  \url{https://shekarpour.github.io/cevo.io/}).
This ontology relies on an abstract conceptualization of English verbs provided by Beth Levin in \cite{levin_english_1993}.
Such an abstract conceptualization largely obviates deficiencies in (i) relation extraction from text, (ii) contextual equivalencing of relations, and (iii) diversity of ontologies.
This ontology presents more than 230 event classes for over 3,000 English verbs as individuals.
We plan to extend CEVO in the direction of integrating and interlinking to other existing ontologies, especially those that contain a conceptualization of events. Currently, we are applying CEVO to the domain of news for extracting events from news headlines.


\bibliographystyle{plain}
\bibliography{bib}

\end{document}